\documentclass[a4paper, 10pt, conference]{ieeeconf}

\newif\ifarwfinalcopy

\arwfinalcopytrue

\IEEEoverridecommandlockouts

\usepackage{textcomp}
\usepackage{graphics} \usepackage{epsfig} \usepackage{mathptmx} \usepackage{times} \usepackage{amsmath} \usepackage{amssymb}  \usepackage{float}
\usepackage{lineno}
\usepackage{tikzpagenodes}
\usepackage{background}
\usepackage{listings}
\usepackage[normalem]{ulem}
\usepackage{caption}
\usepackage{subcaption}
\usepackage{hyperref}
\usepackage{grffile} \usepackage{amsmath}
\usepackage{mathtools}
\usepackage{bm}
\usepackage{booktabs}
\usepackage{multicol}
\usepackage{letltxmacro}

\LetLtxMacro\oldttfamily\ttfamily
\DeclareRobustCommand{\ttfamily}{\oldttfamily\csname ttsize\endcsname}
\newcommand{\setttsize}[1]{\def\ttsize{#1}}

\definecolor{orchid}{rgb}{0.85, 0.44, 0.84}
\definecolor{palecopper}{rgb}{0.85, 0.54, 0.4}
\definecolor{sapgreen}{rgb}{0.31, 0.49, 0.16}
\definecolor{amber(sae/ece)}{rgb}{1.0, 0.49, 0.0}
\definecolor{mediumpersianblue}{rgb}{0.0, 0.4, 0.65}

\newcommand{\mat}[1]{\mathbf{#1}}

\DeclarePairedDelimiter\abs{\lvert}{\rvert}\DeclarePairedDelimiter\norm{\lVert}{\rVert}

\newcommand{\figref}[1]{Fig. \ref{#1}}
\newcommand{\secref}[1]{Sec. \ref{#1}}
\makeatletter
\let\oldabs\abs
\def\abs{\@ifstar{\oldabs}{\oldabs*}}
\let\oldnorm\norm
\def\norm{\@ifstar{\oldnorm}{\oldnorm*}}
\makeatother

\definecolor{mygreen}{rgb}{0,0.3,0}
\definecolor{mygray}{rgb}{0.5,0.5,0.5}
\definecolor{mymauve}{rgb}{0.58,0,0.82}
\definecolor{verylightgray}{rgb}{0.95,0.95,0.95}

\ifarwfinalcopy
\backgroundsetup{color=white}
\else

\linenumbers

\newcommand{\MyARWConfidentialLogo}{\begin{tikzpicture}[remember picture,overlay]
\node[align=center,text=blue] at ([yshift=1em]current page text area.north) {\Large \#\#\# ARW 2023 SUBMISSION: CONFIDENTIAL REVIEW COPY \#\#\#};
\end{tikzpicture}}

\SetBgContents{\MyARWConfidentialLogo}\SetBgPosition{current page.north west}\SetBgOpacity{0.5}\SetBgAngle{0.0}\SetBgScale{1.0}

\fi

\newcommand{\libname}[1]{{\setttsize{\small}\texttt{#1}}}

\title{\LARGE \bf
Differentiable Forward Kinematics for TensorFlow\,2
}

\ifarwfinalcopy
\author{Lukas Mölschl$^{1}$, Jakob Hollenstein$^{2}$,  and Justus Piater$^{3}$
\thanks{$^{1}$Lukas Mölschl is a student in Faculty of Informatics at the TU Wien, Erzherzog-Johann-Platz 1
1040 Vienna, Austria {\tt\small lukas@moelschl.com}}\thanks{$^{2}$Jakob Hollenstein is with the Department of Computer Science, University of Innsbruck, Technikerstraße 21a, 6020 Innsbruck, Austria
{\tt\small 
jakob.hollenstein@uibk.ac.at}}\thanks{$^{3}$Justus Piater is with the Department of Computer Science, University of Innsbruck, Technikerstraße 21a, 6020 Innsbruck, Austria
{\tt\small 
justus.piater@uibk.ac.at}}}
\else
\author{Anonymous, J. D.}  \fi

\makeatletter
\begin{document}

\maketitle

\newcommand{\myparagraph}[1]{\par\textbf{#1}\space}
\makeatletter
\let\oldmyparagraph\myparagraph
\renewcommand\myparagraph[1]{\oldmyparagraph{#1}\@ifnextchar\par{\@gobble}{}}
\makeatother
\begin{abstract}
  Robotic systems are often complex and depend on the integration of a large number of
  software components. 
One important component in robotic systems provides the calculation of \emph{forward kinematics}, which is required by both motion-planning and perception related components. 
End-to-end learning systems based on deep learning require passing gradients across component boundaries.

  Typical software implementations of forward kinematics are not differentiable, and thus prevent the construction of gradient-based, end-to-end learning systems.
  
In this paper we present a library compatible with \emph{ROS}-\emph{URDF} that computes forward kinematics while simultaneously giving access to the gradients w.r.t. joint configurations and model parameters, allowing gradient-based learning and model identification. 
Our Python library is based on Tensorflow~2 and is auto-differentiable. It supports calculating a large number of
  kinematic configurations on the GPU in parallel, yielding a
  considerable performance improvement compared to sequential CPU-based
  calculation.\\ \ifarwfinalcopy
  \url{https://github.com/lumoe/dlkinematics.git}
\else
\url{https://github.com/*/*}
\fi
  
\end{abstract}

\section{INTRODUCTION}

Adaptable, learning robots are an active topic of research. One
promising approach for the realization of such robotic systems are
end-to-end trainable systems. End-to-end trainable means that the
software components of the system are able to adapt and to pass
learning information across component boundaries. Deep-learning-based
end-to-end trainable systems are typically optimized by gradient
descent and the learning information consists of the gradients of a
components outputs with respect to its inputs or
parameters. Components that can provide the gradient information are
called \emph{differentiable}.  Most robotic system component
implementations are not differentiable and thus prevent the
construction of end-to-end trainable systems.

Calculating the \emph{forward kinematics}, the end-effector pose and all intermediate transformation matrices, is a necessary task in virtually all robotic
systems. Although components to calculate forward-kinematics are typically provided by robotic
frameworks, such as the Robot Operating System (ROS)~\cite{quigleyROSOpensourceRobot}, they are usually not differentiable.

An easy to use, differentiable forward kinematics implementation benefits a wide range of tasks.
For example,
James et al. \cite{jamesTransferringEndtoEndVisuomotor2017} resort to learning the
forward kinematics using a deep neural network.  They train a
visuomotor control policy that outputs target positions based on the
camera image and the robots joint positions. This architecture
has to implicitly learn the forward kinematics and could instead
benefit from an explicit forward kinematic component as proposed in this
paper.  Another example is the use of a kinematic model as an inductive bias: Zhou et al.~\cite{zhouDeepKinematicPose2016} propose
using a kinematic layer in vision-to-pose regressions and show
improved performance for both, a toy robot arm example, as well as
full-fledged human pose estimation.  Villegas et al.~\cite{Villegas_2018_CVPR} 
also found that the accuracy of pose estimation can be improved by the use
of a kinematic layer. In contrast to explicitly building the forward-kinematic architecture for each kinematic model by adjusting the source code, our library provides a forward-kinematic layer based on the Universal Robot Description Format (URDF) description of the kinematic model and thus greatly simplifies the implementation.
Calculating inverse kinematics is also often
required and learning the inverse kinematic mapping has been explored~\cite{kokerStudyNeuralNetwork2004a,dukaNeuralNetworkBased2014}. Even such learning approaches can benefit from a differentiable forward-kinematic component, for example by verifying the estimated inverse kinematics against a
known-good forward-kinematic calculation or by refining a predicted inverse mapping: gradient-based local optimization of the forward kinematics can improve the accuracy of the predicted joint configuration.

In this paper we present \libname{DLKinematics}, a building
block for end-to-end deep learning in robotic projects.
Our library allows batch computation of forward kinematics, greatly improving the computational efficiency as shown in our empirical analysis. 
The simplicity of our library---calculation of kinematic chains described in URDF files is possible with two lines of code---further unlocks the potential of deep learning techniques in robotics.
To the best of our knowledge, we present the first Python library based on the widely used deep-learning package Tensorflow~2, that provides auto-differentiable forward kinematic calculations for kinematics described in the Universal Robot Description Format (URDF) popularized by ROS.

\subsection{Related Work}

Robotic software frameworks (e.g.,
ROS~\cite{quigleyROSOpensourceRobot},
armarx~\cite{vahrenkampRobotSoftwareFramework2015}) typically include
components to describe kinematic chains and perform kinematic motion
planning, e.g.~KDL~\cite{herzigHrlkdl2021}, MoveIT~\cite{
  colemanReducingBarrierEntry2014a,gornerMoveItTaskConstructor2019},
which are not differentiable.  This
means they do not provide the gradient information of their outputs
with respect to their inputs or parameters.
This restriction impedes a straightforward application of gradient-based learning techniques.

The difficulty of writing differentiable components for end-to-end
learning has been drastically reduced by the advent and subsequent
wide availability of auto-differentiation libraries, most prominently
for \libname{Python}~\cite{abadiTensorFlowSystemLargescale2016,
  jax2018github, pytorch} but also for languages more traditionally
associated with robotics such as \libname{C++}~\cite{bell2012cppad}.

The availability of automatic differentiation libraries reduces the barrier to 
entry and blurs the line between traditional approaches and deep-learning. For example,
Ledezma et al. \cite{ledezmaFOPNetworksLearning2018,diazledezmaFirstorderprinciplesbasedConstructiveNetwork2017}
proposed the \emph{first-order-principle networks}, that combine
first-order engineering principles with deep-learning models to
calculate robot dynamics. In a similar fashion
Meier et al.\cite{meierDifferentiableLearnableRobot2022} published a
robot-dynamics component that seamlessly integrates with
\libname{pytorch}, while Carpentier et al.~\cite{carpentierPinocchioLibraryFast2019a}
propose a forward- and inverse-dynamics library written in
\libname{C++}. Beyond these methods are the full-fledged
differentiable simulation libraries
\libname{brax}~\cite{brax2021github}, and
\libname{TDS}~\cite{heiden2021neuralsim}. While the former is written
in \libname{jax} and thus seamlessly integrates into the \libname{jax}
and \libname{TensorFlow} frameworks, the latter provides similar
capabilities but does not integrate into the Python learning
systems. Both of these libraries are full-fledged simulators including
simulation of contact dynamics but they are not tailored to extract
the forward kinematics.  The component by Meier et al.~\cite{meierDifferentiableLearnableRobot2022} is the
closest substitute for calculating the forward kinematics in \libname{pytorch}, while our component is
built for \libname{TensorFlow}.
The benefits of our component are the ease of use and integration for calculating forward-kinematics from arbitrary URDF models in Tensorflow.

\section{Method}
\emph{Positions} and \emph{orientations} in 3D space are defined relative to a reference frame, the base frame, where $(x, y, z) \in \mathbb{R}^3$
defines the translations relative to the reference frame and $(\alpha,
\beta, \gamma) \in \mathbb{R}^3$ define the rotations around $(x, y, z)$ relative to the orientation of
the reference frame.  
The combination $(x, y, z, \alpha, \beta, \gamma)$ of position and orientation is
called a \emph{pose} and consists of the translation and the angles describing the rotation. Each pose can
define its own frame of reference, where the pose would be expressed
as the zero pose $(0, 0, 0, 0, 0, 0)$.  Thus a pose can be expressed
as the \emph{transformation} of the zero pose to or from a different reference
frame. In this paper we express coordinates as homogeneous coordinates:
\begin{equation}
  \begin{bmatrix}x & y & z \end{bmatrix}^T \rightarrow \begin{bmatrix}x & y & z & 1\end{bmatrix}^T
\end{equation}
This allows us to express transformations $\mat{T}_{i:j}$ between two
reference frames $i$ and $j$ as a $(4 \times 4)$ matrix, consisting of a
$(3 \times 3)$ rotation matrix $\mat{R}_{i:j}$ and a $(3\times 1)$
translation matrix $\mat{P}_{i:j}$:
\begin{equation}
  \mat{T}_{i:j} = \begin{bmatrix}
    \mat{R}_{i:j} & \mat{P}_{i:j} \\
    0 & 1 \\
  \end{bmatrix}
\end{equation}

The rotation matrix $\mat{R}_{i:j}$ is constructed from rotations around each axis.
\begin{align}
  \mat{R}{_{i:j}} &= \mat{R}{_{z,{i:j}}(\gamma_{i:j})}  \mat{R}{_{y,{i:j}}(\beta_{i:j})}  \mat{R}{_{x,{i:j}}(\alpha_{i:j})} \\
\mat{R}_{x,{i:j}}(\alpha_{i:j}) &= \begin{bmatrix}
1 & 0 & 0 \\
0 & \cos{\alpha_{i:j}} & -\sin{\alpha_{i:j}} \\
1 & \sin{\alpha_{i:j}} & \cos{\alpha_{i:j}} \\
\end{bmatrix}
\\
\mat{R}_{y,{i:j}}(\beta_{i:j}) & = \begin{bmatrix}
\cos\beta_{i:j} & 0 & \sin\beta_{i:j} \\
0 & 1 & 0 \\
-\sin(\beta_{i:j}) & 0 & \cos\beta_{i:j} \\
\end{bmatrix}
\\
\mat{R}_{z,{i:j}}(\gamma_{i:j}) & = \begin{bmatrix}
\cos\gamma_{i:j} & -\sin\gamma_{i:j}  & 0 \\
\sin\gamma_{i:j}  & \cos\gamma_{i:j} & 0 \\
0 & 0 & 1 \\
\end{bmatrix}
\end{align}
The result is a single rotation matrix with order $z, y, x$ shown below. For brevity we denote $\cos(\alpha_{i:j})$ by $c_{\alpha_{i:j}}$ and $\sin(\alpha_{i:j})$
by $s_{\alpha_{i:j}}$. $\mat{R}_{i:j}$:
\begin{align}
\begin{bmatrix}
\label{eq:rotation-complete}
c_{\alpha_{i:j}} c_{\gamma_{i:j}} & -c_{\alpha_{i:j}}  s_{\gamma_{i:j}} + c_{\gamma_{i:j}}  s_{\alpha_{i:j}}  s_{\beta_{i:j}} & c_{\alpha_{i:j}}  c_{\gamma_{i:j}}  s_{\beta_{i:j}} + s_{\alpha_{i:j}} s_{\gamma_{i:j}} \\
c_{\beta_{i:j}} s_{\gamma_{i:j}} & c_{\alpha_{i:j}} c_{\gamma_{i:j}} + s_{\alpha_{i:j}} s_{\beta_{i:j}} s_{\gamma_{i:j}} & c_{\alpha_{i:j}} s_{\beta_{i:j}} s_{\gamma_{i:j}} - c_{\gamma_{i:j}} s_{\alpha_{i:j}} \\
s_{\gamma_{i:j}} & c_{\beta_{i:j}} s_{\alpha_{i:j}} & c_{\alpha_{i:j}} c_{\beta_{i:j}}
\end{bmatrix}
\end{align}
Given multiple transformations $\mat{T}_{i:i+1}, \ldots
\mat{T}_{j-1:j}$, a combined transformation $\mat{T}_{i:j}$ is
calculated by:
\begin{equation}
\mat{T_{i:j}} = \mat{T}_{i:i+1}  \mat{T}_{i+1:i+2}  \dots  \mat{T}_{j-2:j-1}  \mat{T}_{j-1:j}
\end{equation}

\begin{figure}[bt]
    \centering
    \includegraphics[width=\columnwidth]{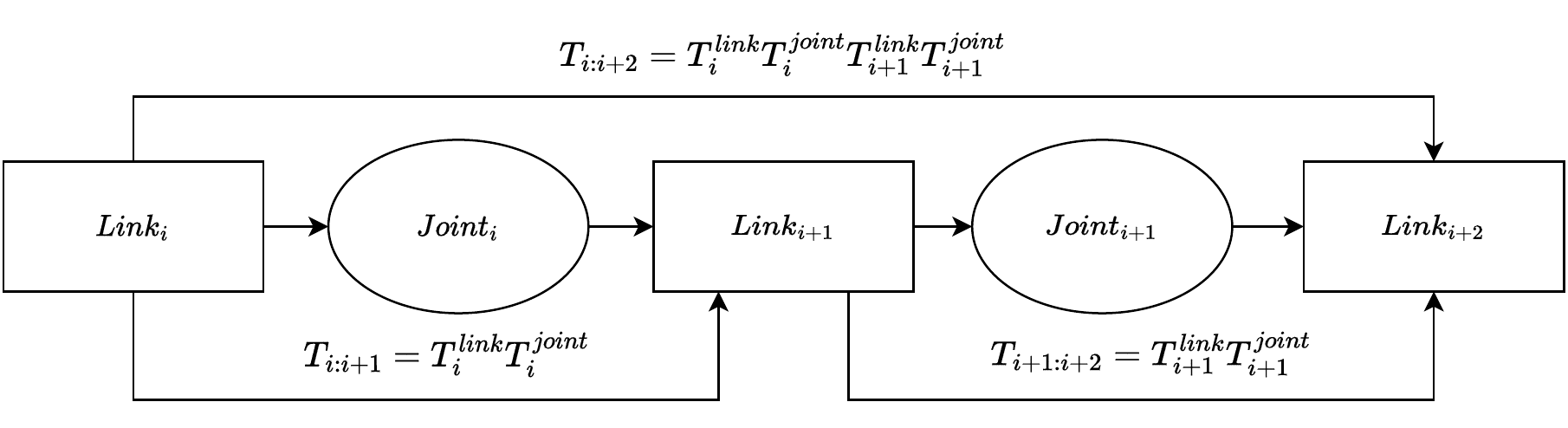}
    \caption{Transformations between $\textrm{link}_n$ to $\textrm{link}_{n+2}$ visualized}
    \label{fig:transformation_chain}
\end{figure}

In this paper we describe the kinematics of a robot as a tree
consisting of links (static) and joints (moveable). Each node in the
tree defines a reference frame, for example the frame at the end of a
link, or the frame describing the displacement or rotation due to a
joint. Two links are always connected by a joint (\figref{fig:transformation_chain}).

A \emph{kinematic chain} describes a sub-path between nodes in this
tree, e.g. $\textrm{link}_i$ \ldots $\textrm{link}_j$. Using the
transformations along the path $\mat{T}_{i:i+1}, \ldots,
\mat{T}_{j-1:j}$, the complete transformation $\mat{T}_{i:j}$ from
$\textrm{link}_i$ to $\textrm{link}_j$ is calculated.  For example, in
a robot with two end-effectors, the two transformations between the
robot base and each of the end-effectors are described by two separate
kinematic chains.
\subsection{Implementation Details}
\subsubsection{Unified Robot Description Format (URDF)} URDF is a standardized
XML file format used to describe the kinematic properties of a robot,
i.e. its joints and links. The URDF \cite{UrdfXMLJoint} specifies
different types of joints: revolute (hinge), continuous (continuous
hinge), prismatic (sliding), fixed, planar and floating
(6-DoF joint). The most generic joint is the 6-DoF joint (floating)
joint. According to the URDF specifications, a constraint exists
wherein each joint must have exactly one parent link and child link. 
It is not possible for a joint to be followed by another joint or for a 
link to be followed by another link.
 The URDF defines the kinematic structure as a tree with
parent-child relationships and defines relative translations and
rotations. Joint properties such as limits can be specified. For
further details refer to \cite{UrdfXMLJoint}.

\subsubsection{Modelling by 6-DoF joint\label{sec:universal_joint}} The 6-DoF joint fully parameterizes a $(4\times 4)$ homogeneous transformation
matrix.  The 6-DoF joint is parameterized by six parameters in the
form $[x,y,z,\alpha, \beta, \gamma]$ where $[x, y, z]$ represent
translation and $[\alpha, \beta, \gamma]$ describes rotation around
each axis. All other joint types can be derived from this joint by
fixing all but the relevant axes.  For example, a prismatic joint,
with one degree of freedom, that can slide along the $y$ axis, can be
represented as a 6-DoF joint in the form $[0,y,0,0,0,0]$. A revolute joint, with one degree of freedom, that rotates around the $z$ axis, would be parameterized by $\gamma$ in the form $[0,0,0,0,0,\gamma]$.
This structure is used to model all joint types defined in the URDF
specifications.

\subsubsection{Vectorized Implementation} Let a kinematic chain of interest be denoted by
 $\mat{T}^L_{0:1}, \mat{T}^J_{0:1} \ldots \mat{T}^L_{n-1:n}\mat{T}^J_{n-1:n}$, and
$\mat{T}_{0:n}$ denotes the complete transformation from the initial frame to the end
of the kinematic chain, e.g. the end-effector.
The transformation $\mat{T}_{0:n}$ is composed alternatingly of two components: a transformation along a link $\mat{T}^L_{i:i+1}$, and a transformation along a joint $\mat{T}^J_{i:i+1}$. Since the parameters of the links are static, $\mat{T}^L_{i:i+1}$ are pre-computed during the initialization phase according to the URDF, while $\mat{T}^J_{i:i+1}$ are paremeterized.

Let $b$ denote the batch-size, i.e. the number of different joint configurations for which the
forward kinematic calculations are performed, and $n$ denotes the number of joints.

The dynamic transformations $\mat{T}^J_{i:i+1}$ along the joints are calculated by partially parameterizing
6-DoF joints. These parameterizations are stored in a $(b \times n
\times 6)$ tensor $\mat{Q}$:
\begin{gather}
  \mat{Q} = \begin{bmatrix}
    \begin{bmatrix}
      x^{(1)}_{0:1} & y^{(1)}_{0:1} & z^{(1)}_{0:1} & \alpha^{(1)}_{0:1} & \beta^{(1)}_{0:1} & \gamma^{(1)}_{0:1} \\
      \vdots & \vdots & \vdots & \vdots & \vdots & \vdots  \\
      x^{(1)}_{n-1:n} & y^{(1)}_{n-1:n} & z^{(1)}_{n-1:n} & \alpha^{(1)}_{n-1:n} & \beta^{(1)}_{n-1:n} & \gamma^{(1)}_{n-1:n}
    \end{bmatrix}
    \\
    \vdots
    \\
    \begin{bmatrix}
      x^{(b)}_{0:1} & y^{(b)}_{0:1} & z^{(b)}_{0:1} & \alpha^{(b)}_{0:1} & \beta^{(b)}_{0:1} & \gamma^{(b)}_{0:1} \\
      \vdots & \vdots & \vdots & \vdots & \vdots & \vdots  \\
      x^{(b)}_{n-1:n} & y^{(b)}_{n-1:n} & z^{(b)}_{n-1:n} & \alpha^{(b)}_{n-1:n} & \beta^{(b)}_{n-1:n} & \gamma^{(b)}_{n-1:n}
    \end{bmatrix}
  \end{bmatrix}
\end{gather}
The parameterization of the degrees of freedom $\theta^{(k)}_1, \ldots \theta^{(k)}_m$ are modelled,
following the explanation in \secref{sec:universal_joint} by modifying the corresponding parameters in $\mat{Q}$. 
How to map from $\theta^{(k)}_i$ to parameters of $\mat{Q}$
is determined by indexing.
The batch of joint configurations, $\bm{\Theta}$ is a $(b m \times 1)$ matrix:
\begin{equation}
\bm{\Theta} = \begin{bmatrix}
\theta^{(1)}_1 & \ldots & \theta^{(1)}_m \ldots \theta^{(b)}_1 & \ldots & \theta^{(b)}_m \ldots
\end{bmatrix}^\top
\end{equation}
The index matrix $\mat{P}$, is a $(b m \times 3)$ matrix:
\begin{equation}
    \mat{P} = \begin{bmatrix}
       c^{(1)}_1 & d^{(1)}_1 & e^{(1)}_1 \\
       \vdots \\
       c^{(1)}_m & d^{(1)}_m & e^{(1)}_m \\
       \vdots \\
       c^{(b)}_1 & d^{(b)}_1 & e^{(b)}_1 \\
       \vdots \\
       c^{(b)}_m & d^{(b)}_m & e^{(b)}_m \\
    \end{bmatrix}
\end{equation}

$c^{(i)}_j$ is set to $i$ and maps $\theta^{(i)}_j$ to the $i$-th batch in $\mat{Q}$, $d^{(i)}_j \in \{1, \ldots n\}$ denotes the row of the $i$-th batch in $\mat{Q}$. 
$e^{(i)}_j$ denotes the target parameter $ \{1, \ldots, 6\} \mapsto \{x, \ldots \gamma\}$.
These indices are
only calculated during the first call, allowing for sparse updates to
the zero matrix during the forward call and the incorporation of
dynamic joint configurations into the matrix.
Mapping from $\bm{\Theta}$ to $\mat{Q}$ is efficiently implemented using \texttt{scatter\_nd} and is executed on the GPU.

For all 6-DoF joint configurations specified in $\mat{Q}$ the
associated transformation matrices $\mat{T}^{J(k)}_{i:j}$ are calculated in
parallel, yielding a $(b \times n \times 4 \times 4)$ tensor,
$\mat{\mathfrak{T}}^J$.
\newcommand{\bmat}[1]{\begin{bmatrix} {#1} \end{bmatrix}}
\begin{equation}
  \mathfrak{T}^J = \begin{bmatrix}
    \bmat{\mat{T}^{J(1)}_{0:1}} & \ldots & \bmat{\mat{T}^{J(1)}_{n-1:n}} \\
    \vdots & & \vdots \\
    \bmat{\mat{T}^{J(b)}_{0:1}} & \ldots & \bmat{\mat{T}^{J(b)}_{n-1:n}} \\
    \end{bmatrix}
\end{equation}
The pre-computed static transformations $\mat{T}^{L}_{i:i+j}$ are stored as a $(b \times n \times 4 \times 4)$ tensor $\mathfrak{T}^L$ (links):
\newcommand{\bbmat}[1]{\begin{bmatrix} {#1} \end{bmatrix}}
\begin{gather}
  \mathfrak{T}^L = \begin{bmatrix}
    \bbmat{\mat{T}^{L(1)}_{0:1}} & \ldots & \bbmat{\mat{T}^{L(1)}_{n-1:n}} \\
    \vdots & & \vdots \\
    \bbmat{\mat{T}^{L(b)}_{0:1}} & \ldots & \bbmat{\mat{T}^{L(b)}_{n-1:n}} \\
    \end{bmatrix}\end{gather}
A combined transformation $\mathfrak{T}^{LJ}$ is  calculated by combining all $k,i$: $\mat{T}^{LJ(k)}_{i:i+1} =\mat{T}^{L(k)}_{i:i+1}\mat{T}^{J(k)}_{i:i+1}$. This operation is efficiently parallelized across the batch dimension $k\in \{1,\ldots b\}$ and link-joint segment $i \in \{1,\ldots,n\}$.

Following \eqref{eq:rotation-complete}, efficiently implemented using a
\texttt{scan} operation, the $b$ kinematic chains of transformations of $\mathfrak{T}^{LJ}$
are sequentially combined into complete forward transformations $\mathfrak{T}'$ :
\begin{equation}
  \mathfrak{T}' = \begin{bmatrix}
    \bmat{\mat{T}^{LJ(1)}_{0:1}} & \ldots & \bmat{\mat{T}^{LJ(1)}_{0:n}} \\
    \vdots & & \vdots \\
    \bmat{\mat{T}^{LJ(b)}_{0:1}} & \ldots & \bmat{\mat{T}^{LJ(b)}_{0:n}} \\
    \end{bmatrix}
\end{equation}
Depending on the use case $\mathfrak{T}'$ or simple the final transformations are returned, i.e. $(b \times 4 \times 4)$:
\begin{equation}
\begin{bmatrix}
    \bbmat{\mat{T}^{LJ(1)}_{0:n}} &
    \ldots &
    \bbmat{\mat{T}^{LJ(b)}_{0:n} }
\end{bmatrix}
\end{equation}
\begin{figure}[bt]
  \begin{subfigure}[c]{\columnwidth}
    \centering \includegraphics[trim={0 0 0
      0},clip,width=0.9\columnwidth]{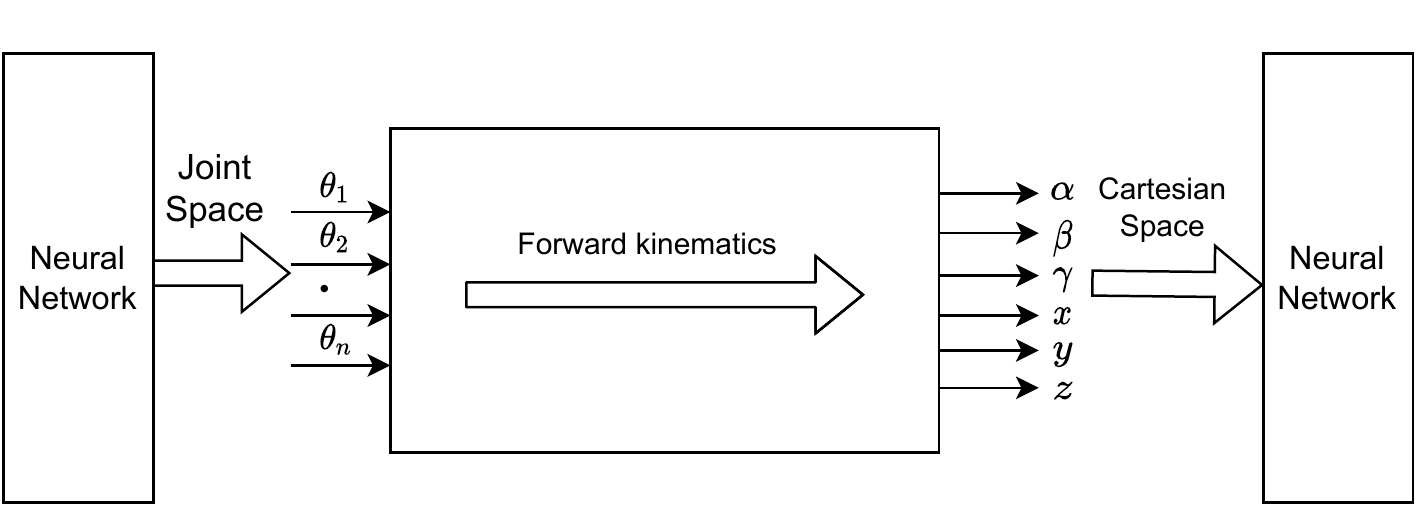}
    \subcaption{The Forward Kinematics calculation can be used as a
      building block in any algorithm but can also easily be embedded
      as a neural network layer.}
  \label{fig:embed_keras_illustration}
  \end{subfigure}
  \centering
  \begin{subfigure}[c]{.9\columnwidth}
    \noindent
    \begin{lstlisting}[caption={},language=python]
from dlkinematics.training_utils import ForwardKinematics

model = keras.Sequential()

FK_layer = ForwardKinematics(
urdf_file = 'path/to/urdf',
base_link = 'link0',
end_link = 'linkN',
batch_size = 32)

# model.add(... NN layers ...)
model.add(FK_layer)
# model.add(... NN layers ...)
\end{lstlisting}
\subcaption{Embedding the calculation as a layer requires two lines of code.}
  \label{fig:embed_keras_code}

  \end{subfigure}
  \caption{Example of embedding FK calculations for a URDF file as a neural network layer.}
\end{figure}

\section{Library Usage Examples}
\subsection{Embedding Forward Kinematics in a Neural Network}

Keras is a high-level API for TensorFlow. It allows users to easily
define and train deep neural networks by specifying the network in
terms of layers, thereby abstracting the underlying workings of
TensorFlow.

\libname{DLKinematics} integrates into this ecosystem by providing a
class that can be added as a Keras layer, illustrated in
Fig.~\ref{fig:embed_keras_illustration}. Creating such a
forward kinematics layer and embedding it into a Keras model requires
just two lines of code as in
Fig.~\ref{fig:embed_keras_code}. In such a setup, the network would
calculate joint configurations, e.g., based on an image, and the
forward kinematics layer would output the pose. This pose can
be used as part of the loss function or be further processed by 
additional neural-network layers.

\subsection{Calculating Jacobian Matrix}
A typical requirement in robotics is to calculate the
Jacobian of the end-effector pose with respect to the joints,
which is typically used to capture the relationship between joint
velocities and the end-effector velocities.

\begin{figure}[bt]
  \centering
\noindent
\begin{lstlisting}[caption={},language=python]
import tensorflow as tf
from dlkinematics.urdf import chain_from_urdf_file
from dlkinematics.dlkinematics import DLKinematics
from dlkinematics.tf_transformations import pose_from_matrix

# Load URDF
chain = chain_from_urdf_file('data/human.urdf')

# Create DLKinematics
dlkinematics = DLKinematics(
    chain,
    base_link="link0",
    end_link="link3",
    batch_size=2)

# Joint configuartion
thetas = tf.convert_to_tensor([[1., 2., 3.], [3., 4., 6.]], dtype=tf.float32)

# Forward pass
with tf.GradientTape(persistent=True) as tape:
    tape.watch(thetas)
    fk_res = dlkinematics.forward(tf.reshape(thetas, (-1,)))
    p = pose_from_matrix(fk_res)

jacobian = tape.batch_jacobian(p, thetas)
\end{lstlisting}
\caption{Example of computing the Jacobian for a batch of sample joint configurations.}
\label{list:jacobian}
\end{figure}

The computation of the Jacobian matrix with DLKinematics
is straightforward, as depicted in Fig. ~\ref{list:jacobian} 
In this listing, the forward kinematics for a batch of joint
configurations and the resulting Jacobian matrix of the position and
rotation of the end-effector with respect to the joint configuration
are computed.

\begin{figure*}[tb] 
  \begin{multicols*}{2}
  \begin{minipage}{.99\columnwidth}
  \centering
  \begin{subfigure}[c]{.7\columnwidth}
    \centering \includegraphics[trim={0 0 0.6cm
      0},clip,width=0.9\columnwidth]{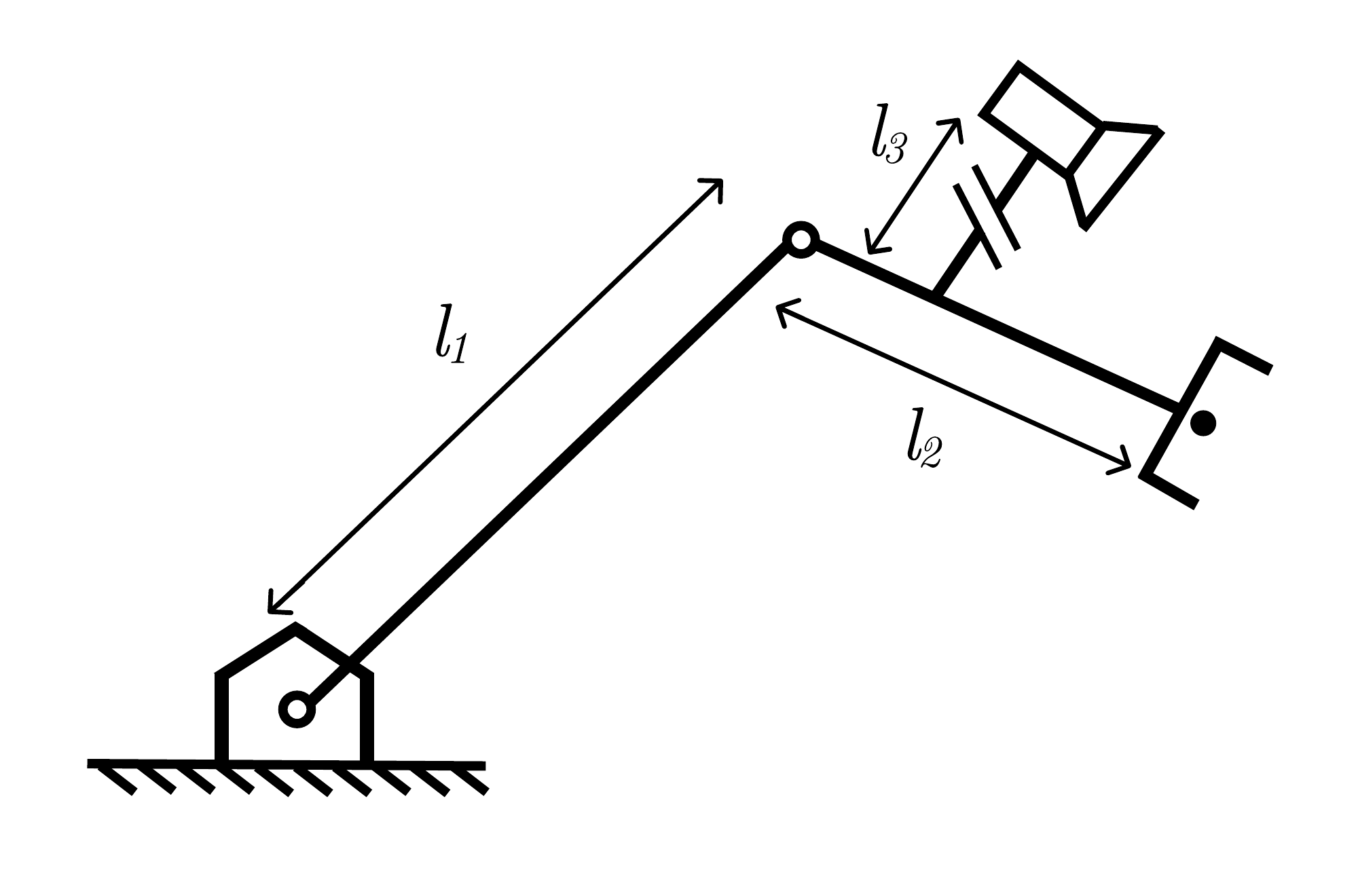}
    \subcaption{Kinematic model identification: The length and orientation of $l_3$ is unknown or
      inaccurate with respect to the end-effector and needs to be estimated.}
    \label{fig:kinematic_identification_robot}
  \end{subfigure}
  \begin{subfigure}[c]{.9\columnwidth}
    \centering \includegraphics[trim={0 0 0.6cm
      0},clip,width=0.9\columnwidth]{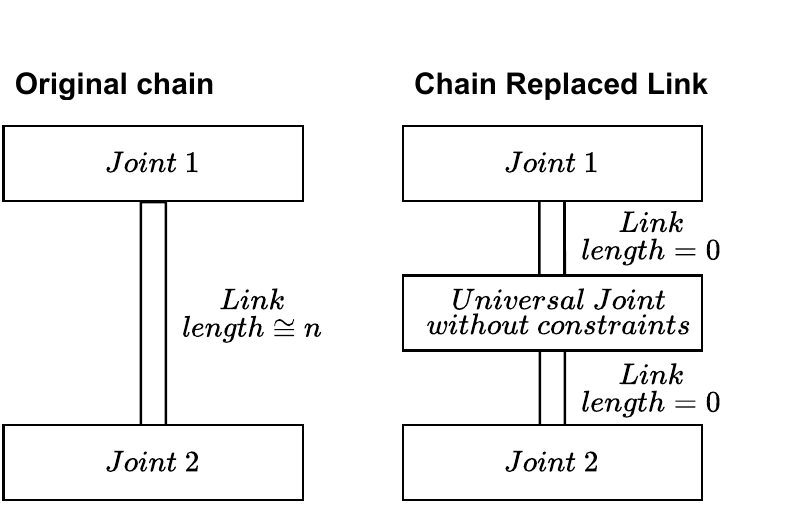}
    \subcaption{\libname{DLKinematics} provides a function that
      replaces a link with a joint with six degrees of freedom
}
    \label{fig:kinematic_identification_replace_link}
  \end{subfigure}
\end{minipage}
\begin{minipage}{0.99\columnwidth}
\centering
  \begin{subfigure}[c]{.9\columnwidth}
    \noindent
    \begin{lstlisting}[caption={},language=python]
from dlkinematics.dlkinematics import DLKinematics
from dlkinematics.training_utils import subsitute_link_with_joint
from evaluation.estimate_model_params import ParamEstimator
from evaluation.generate_dataset import Generator

# Loading kinematic chain ..

# Replace link with 6-DoF joint
new_chain = subsitute_link_with_joint(urdf=dl_kin.urdf,
        target_link=target_link,root=root, last=last)
dl_kin_new = DLKinematics(new_chain, root, last, batch_size=batch_size)

sample_generator = Generator(urdf_file=path_to_urdf,
    root_link=root, end_link=last, batch_size=batch_size)

estimator = ParamEstimator(dl_kin, dl_kin_new, batch_size=batch_size)

# Optimization loop
for idx, q_sample in enumerate(sample_generator.joint_samples()):
    res, iterations = estimator.step(q_sample.flatten())
    if np.mod(idx, num_samples) == 0:
        break

print(f'{np.round(estimator.qs.numpy(), 3)}')
\end{lstlisting}
\subcaption{}
\label{fig:kinematic_identification_code}
  \end{subfigure}
  \end{minipage}
  \caption{Example of identifying link parameters through an optimization method.}
\end{multicols*}
\end{figure*}

\subsection{Kinematic Model Identification}
The kinematics description of robots and its physical realization
might differ due to inaccuracies in production or the setup of the
robot, for example the placement of cameras acting as visual
sensors. An example of this is illustrated illustrated in
Fig.~\ref{fig:kinematic_identification_robot}, where the link
describing the camera position with respect to the end-effector is
unknown or inaccurate. One way to overcome such a problem could be to
record the location of the end-effector of a physical robot for a set
of joint configurations. This data can then be used to compute the
error between the estimated and expected location of the end-effector.
In the next step our library can easily be used to iteratively
estimate the real length and orientation of the links. We achieve this
by replacing a link with a 6-DoF joint that gets initialized with
the parameters of the original link from the URDF file (see
Fig.~\ref{fig:kinematic_identification_replace_link}).  A 6-DoF joint
has six degrees of freedom and is therefore not bound to the
original translation and orientation of the link. Using a
gradient-based algorithm, estimate all six parameters of the 6-DoF joint.
In addition to that, the \libname{DLKinematics} provides
functions to replace links with a 6-DoF joint
(Fig.~\ref{fig:kinematic_identification_code}).  In an experiment with
a chain of length four, we were able to estimate all six parameters of
a link using gradient descent. Using a single fixed joint
configuration and providing no prior estimate of the link,
i.e. initializing the translation and rotation to zero, the algorithm
converged in under $30$ seconds (under $3000$ gradient steps) with an
accuracy in the range of $10^{-4}$ for each axis.

\section{Experiment: Batch Processing}

\begin{table}[b]
  \centering
  \begin{tabular}{@{}lrrrr@{}}
    \toprule
    Batch Size                   & 1      & 256     & 1024    & 4096    \\
    \midrule
    pykdl {[}ops/s{]}            & 20,010 & -       & -       & -       \\ 
    DLKinematics CPU {[}ops/s{]} & 485    & 91,980  & 202,819 & 374,491 \\ 
    DLKinematics GPU {[}ops/s{]} & 537    & 114,551 & 338,773 & 802,318 \\
    \bottomrule
  \end{tabular}
  \caption{Batch processing: Speedup in parallelization.}
  \label{table:batch_timings}
\end{table}

In the process of training deep neural networks, samples are usually
passed through the network in batches. 
By utilizing parallel processing wherever realizable, it is possible to simultaneously compute the forward kinematics for a large number of joint configurations. This ability is crucial for 
population-based search methods, which are used, for example, in
trajectory optimization.
In this experiment, we use a kinematic chain with four degrees of freedom and
measure the amount of forward kinematic computations per
second. This is done using the comparison ground-truth library,
\libname{pykdl} on CPU, and our implementation on both CPU (AMD Ryzen 5 3600 6-Core) as well as
GPU (NVIDIA GeForce RTX 3060 Ti). Note that \libname{pykdl}, does not support parallel
calculations. The experiment results on commodity hardware are shown
in Table \ref{table:batch_timings}. Our empirical evaluation shows that for a batch size of one, the
additional call overhead is dominant, our implementation already
achieves a much higher throughput at $256$ parallel computations on
both, the CPU and the GPU.

\section{SUMMARY AND OUTLOOK}

In this paper we present a differentiable forward kinematic library for
kinematic representations described in the ROS URDF format. Our
library is built on and seamlessly integrates with
\libname{TensorFlow} and can be used as a building block in end-to-end
learning systems in just three lines of code.  We highlight the
simplicity of embedding the forward kinematic calculation as a layer
in \libname{Keras} models as well as embedding it in
\libname{TensorFLow} calculations in general, show how to calculate
end-effector jacobians and use it to identify and optimize unknown or
inaccurate kinematic descriptions.

We benchmark our forward kinematic calculation library on a four degree of
freedom kinematic chain, on both CPU and GPU,
comparing the results with \libname{pykdl}, a CPU-based library. We find that large batch computations suffer less from the call overhead associated with \libname{TensorFlow} calculations. 
However, the number of forward-kinematic-calculations per second of
our implementation largely outperforms the \libname{pykdl} CPU-based
library (using batched calculations with a batch size of $1024$
configurations).
The simplicity of our module, combined with its computational efficiency, unravels the application of modern deep learning techniques to robotics, and the inclusion of forward kinematics in end-to-end learning.
\section{Appendix}

\subsection{Rotation Distance Functions}

\begin{figure}[bt]
  \centering
\noindent
\begin{lstlisting}[caption={},language=python]
from dlkinematics.training_utils import phi5_loss

result = dlkinematics.forward([1., 2., 3.])
# target ... transformation matrix [batch size, 4, 4]
error = phi5_loss(result, target)
\end{lstlisting}
\caption{Example use of the rotational-distance metric $\Phi_5$.}
\label{fig:rotation_metrics_code}
\end{figure}

Determining the error between the expected and actual output of a
model is a task often required in machine learning systems. For the
translation it is sufficient to use the Euclidean distance. However,
rotation matrices require different metrics.

\libname{DLKinematics} therefore implements five different distance
functions from~\cite{huynhMetrics3DRotations2009} to compute metrics
for the difference between two rotation matrices.  All functions allow
batch processing of $[batch\_size \times 4 \times 4]$ homogeneous
transformation matrices. The five functions $\Phi_1$, $\Phi_2$,
$\Phi_3$, $\Phi_4$, $\Phi_5$ are briefly described with respect to
their calculation, signature, function name and an example usage is
shown in Fig~\ref{fig:rotation_metrics_code}:

\emph{Euclidean distance of Euler angles} \libname{rotation\_with\_rmse}\\ Computing the Euclidean distance
  between two Euler angles, $\Phi_1 = \sqrt{(\alpha - \hat{\alpha})^2 + (\beta - \hat{\beta})^2 + (\gamma - \hat{\gamma})^2}$.  \\ 
  The function internally extracts the
  Euler angles from the matrix.  Signature:
  $T_1 \times T_2 \rightarrow \mathbb{R}^+$.

  \emph{Euclidean distance between two quaternions} \libname{phi2\_loss}\\ Computing the Euclidean distance between
  two quaternions, $\Phi_2 = \min\{\norm{q - \hat{q}}, \norm{q + \hat{q}}\}$.  
  The function internally extracts quaternions from
  the matrix.  Signature:
  $T_1 \times T_2 \rightarrow \mathbb{R}^+ \in [0, \sqrt{2}]$.

 \emph{Inner products between two quaternions} \libname{phi3\_loss}\\ Computing the inner product of two
  quaternions. $\Phi_3 = \arccos{(\abs{q - \hat{q}})}$ 
  The function internally extracts quaternions from the
  matrix.  Signature:
  $T_1 \times T_2 \rightarrow \mathbb{R}^+ \in [0, \pi / 2]$.

  \emph{Absolute quaternion inner product} \libname{phi4\_loss}\\
  Similar to \libname{phi3\_loss}, except
  that the inverse cosine function is replaced, $\Phi_4 = 1 - (\abs{q - \hat{q}})$. The function
  internally extracts quaternions from the matrix.  Signature:
  $T_1 \times T_2 \rightarrow \mathbb{R}^+ \in [0, 1]$.

\emph{Deviation from the identity matrix} \libname{phi5\_loss}\\ Computes the deviation from the identity
  matrix, $\Phi_5 = \norm{\mat{I} - \mat{R}\mat{\hat{R}^T}}$. Signature:
  $T_1 \times T_2 \rightarrow \mathbb{R}^+ \in [0, 2\sqrt{2}]$.

\section*{ACKNOWLEDGMENT}

We would like to thank our colleagues 
\ifarwfinalcopy
Samuele Tosatto, Erwan Renaudo and Hector Villeda 
\else
Colleague1, Colleague2 and Colleague3
\fi
for their helpful comments towards improving this paper.

{\small
  \bibliographystyle{IEEEtranS}
\bibliography{dl-kinematics,manual}
}

\end{document}